\documentclass[11pt,a4paper]{article}
\usepackage[hyperref]{emnlp2020}
\usepackage{times}
\usepackage{graphicx}
\usepackage{multirow}
\usepackage{makecell}
\usepackage{amssymb}
\usepackage{amsmath}
\usepackage{paralist}

\aclfinalcopy 

\title{A Multi-Task Incremental Learning Framework with Category Name Embedding for 
	Aspect-Category Sentiment Analysis}

\author{
    Zehui Dai*, Cheng Peng, Huajie Chen, Yadong Ding\\
    NLP Group, Gridsum\\
    \texttt{\{daizehui,pengcheng01,chenhuajie,dingyadong\}@gridsum.com}
\\}

\date{}

\begin{document}
\maketitle
\begin{abstract}
(T)ACSA tasks, including aspect-category sentiment analysis (ACSA) and 
targeted 
aspect-category sentiment analysis (TACSA), aims at identifying sentiment 
polarity on predefined categories. Incremental learning on new categories is necessary for (T)ACSA real applications. Though current multi-task learning models achieve good performance in (T)ACSA tasks, they suffer from catastrophic forgetting problems in (T)ACSA incremental learning tasks.
In this paper, to make multi-task learning feasible for incremental learning, we proposed Category Name 
Embedding network 
(\textbf{CNE}-net). We set both encoder and decoder shared among all categories to weaken the catastrophic forgetting problem. Besides the origin input sentence, we applied another input feature, i.e., category name, for task discrimination.
 Our model achieved state-of-the-art 
on two (T)ACSA benchmark datasets. Furthermore, we proposed 
a dataset for (T)ACSA incremental learning and achieved the best performance
compared with other strong baselines.
\end{abstract}

\section{Introduction}
\label{Introduction}

Sentiment analysis has become an increasingly popular natural language 
processing (NLP) task in academia and industry. 
It provides real-time 
feedback on consumer experience and their needs, which helps 
producers to offer better services. 
To deal with the presence of 
multiple categories in one document, 
\textbf{(T)ACSA}
tasks, including aspect-category 
sentiment analysis (\textbf{ACSA}) and targeted aspect-category sentiment analysis (\textbf{TACSA}), were introduced. 

The main purpose for ACSA task 
is to identify sentiment polarity (i.e. positive, neutral, negative and 
none) of an input sentence upon specific predefined categories \cite{DBLP:conf/semeval/MohammadBSK18,DBLP:conf/semeval/WuWLYWH18}. For 
example, as shown in Table \ref{tab:table1}, giving an input sentence ``Food is 
always fresh and hot-ready to eat, but it is too expensive." and predefined categories \{\textit{food}, \textit{service,} \textit{price}, 
\textit{ambience} and \textit{anecdotes/miscellaneous}\}, 
the sentiment of category \textit{food} is positive, the polarity 
regarding to category \textit{price} is negative, while is none for others. 
In this task, the models should 
capture both explicit expressions and implicit expressions. For example, the phrase ``too expensive" indicates the 
negative polarity  in the \textit{price} category, without a direct
indication of ``price".

In order to 
deal with ACSA with both multiple categories and multiple targets,
TACSA task was introduced \cite{DBLP:conf/coling/SaeidiBLR16} to analyze sentiment polarity on a set of predefined target-category pairs. An example is shown in Table \ref{tab:table1},
given targets ``restaurant-1" and ``restaurant-2", in the case ``I like 
restaurant-1 because it's cheap, but restaurant-2 is too 
expansive", the category \textit{price} for target ``restaurant-1" is positive, but is 
negative for target ``restaurant-2", while is none for other target-category pairs. A mathematical definition for (T)ACSA is given 
as follows: giving a 
sentence $s$ as input, a predefined set of targets $T$ and a predefined set of 
aspect categories $A$, a model predicts the sentiment polarity $y$ for 
each target-category pair $\{(t, a) : t \in T, a \in A\}$. For ACSA 
task, there is only one target $t$ in all $(t, a)$ categories. In this paper, in order to simplify the expression in TACSA,
we use predefined categories, which is short for predefined target-category pairs.

\begin{table*}[!t]
	\centering
	\scalebox{0.9}
	{
		\begin{tabular}{|c|c|c|}
			\hline
			Task & Sentence & Labels\\
			\cline{1-3}
			ACSA & \makecell{Food is always fresh and hot-ready to eat, \\but it is 
				too expensive}& \makecell{(food,positive),\\(service, none),\\
				(price, negative), \\(ambience, none)\\(anecdotes/miscellaneous, 
				none)}\\
			\cline{1-3}
			TACSA & \makecell{I like restaurant-1 because it’s cheap, \\but 
				restaurant-2 is too expansive.}& \makecell{(restaurant-1-general, 
				none),\\(restaurant-1-price,positive),\\(restaurant-1-location, 
				none),\\(restaurant-1-safety,none),\\(restaurant-2-general, 
				none),\\(restaurant-2-price,negative),\\(restaurant-2-location, 
				none),\\(restaurant-2-safety,none)}\\
			\cline{1-3}
		\end{tabular}
	}
	\caption{\label{tab:table1} Example and gold standard for (T)ACSA examples.}
\end{table*}

Multi-task learning, with shared encoders but individual decoders for each category, is an approach to analyze all the categories in one sample simultaneously for (T)ACSA \cite{DBLP:journals/corr/abs-1808-01216,DBLP:conf/emnlp/SchmittSSR18}. Compared with single-task ways \cite{DBLP:conf/acl/LiangDXLH19}, multi-task approaches utilize category-specific knowledge in training signals from each task and get better performance. However, current multi-task models still suffer from a lack of 
features such as category name \cite{DBLP:conf/wassa/MeisheriK18}. Models with category name features encoded in the model may further improve the performance.

On the other hand, the predefined categories in (T)ACSA task make the application 
in new categories inflexible, as for (T)ACSA applications, the number of categories maybe 
varied over time. 
For example, \textit{fuel consumption, price level, engine power, space} and so 
on are \textbf{source categories} to be analyzed in the gasoline automotive domain. For 
electromotive domain, source categories in the automotive domain will still be used, while new \textbf{target category} such as
\textit{battery duration} should also be analyzed. 
Incremental learning is a way to solve this problem. Therefore, it is necessary to propose an 
incremental learning task and an incremental learning model concerned with new 
category for (T)ACSA tasks.

Unfortunately, in the current multi-task learning (T)ACSA models, the encoder is shared but the decoders for each category are individual. This parameter sharing mechanism results in only the shared encoder 
and target-category-related decoders are finetuned during the finetuning process, while the decoder of source categories remains unchanged. The finetuned encoder and original decoder of source categories may
cause catastrophic forgetting problem in the origin 
categories. For real applications, high accuracy is excepted in source 
categories and target 
categories. 
Based on the previous researches that decoders between different tasks are usually modeled by mean regularization \cite{DBLP:conf/kdd/EvgeniouP04} 
, an idea comes up to further make the decoders the same by sharing the decoders in all categories to decrease the catastrophic forgetting problem. But here raises another question, how to identify each category in the encoder and decoder shared network? In our approach, we 
solve the category discrimination problem by the input category name feature.

In this paper, 
we proposed a multi-task category name embedding network 
(\textbf{CNE}-net). 
The multi-task learning 
framework makes full use of training signals from all categories. To make it feasible for incremental learning, both encoder and decoders for each category are shared. The category names were applied as another input feature for task discrimination. We also present a new task for (T)ACSA incremental learning. In particular, 
our contribution is three-folded: 

(1) We proposed a multi-task \textbf{CNE}-net framework with both encoder and decoder shared to weaken catastrophic forgetting problem in multi-task learning (T)ACSA model.  

(2) We achieved 
state-of-the-art on the two (T)ACSA datasets, SemEval14-Task4 
and Sentihood.

(3) We proposed a new task for incremental learning in (T)ACSA. By sharing both encoder layers and decoder layers of all the tasks, we  
achieved better results compared with other baselines both in source 
categories and in the target category.

\section{Related Work}
\label{Related Work}

\subsection{Aspect-category Sentiment Analysis}
(T)ACSA task is to 
predict sentiment polarity on a set of predefined categories. 
It is able to 
analyze sentiment in an end-to-end way with explicit expressions or implicit expressions \cite{DBLP:conf/semeval/MohammadBSK18,DBLP:conf/semeval/WuWLYWH18}. 
The earliest works most concerned on feature engineering 
\cite{DBLP:conf/ijcnlp/ZirnNSS11,DBLP:conf/wassa/Wiebe12,DBLP:conf/semeval/WagnerACBBFT14}.
Subsequently, 
\newcite{DBLP:conf/emnlp/NguyenS15,DBLP:conf/eacl/LiakataWZP17,DBLP:conf/wassa/MeisheriK18}
applied neural network models to achieve higher accuracy. 
\newcite{DBLP:conf/aaai/MaPC18} then involved
commonsense knowledge as additional features. 
The 
current approaches consist of multi-task models 
\cite{DBLP:journals/corr/abs-1808-01216,DBLP:conf/emnlp/SchmittSSR18}, 
which analyze all the categories simultaneously in one sample to make 
full use of all the features and labels in the training sample, and single-task 
models that treat one category in one sample 
\cite{DBLP:conf/emnlp/JiangCXAY19}.

\subsection{Multi-Task Learning}
Multi-task learning(MTL) utilizes all the related tasks by 
sharing the commonalities while learning individual features for each sub-task. 
MTL has been
proven to be effective in many NLP tasks, such as 
information retrieval \cite{DBLP:conf/naacl/LiuGHDDW15}, machine translation 
\cite{DBLP:conf/acl/DongWHYW15}, and semantic role labeling 
\cite{DBLP:conf/icml/CollobertW08}. 
For ACSA task, \newcite{DBLP:conf/emnlp/SchmittSSR18} 
applied MTL framework with a shared LSTM encoder and individual decoder classifiers for each category. The multiple aspects in MTL were handled by constrained attention networks with orthogonal and sparse 
regularization \cite{DBLP:conf/emnlp/HuZZCSCS19}.

\subsection{Incremental Learning}
Incremental learning was inspired by adding new abilities to a model without having to retrain the entire model. For 
example, \newcite{DBLP:conf/icmla/DoanK16} presented several random forest models to 
perform sentiment analysis on customers' reviews. Many domain 
adaptation approaches utilizing transfer learning suffer from ``catastrophic 
forgetting" problem \cite{DBLP:journals/neco/FrenchC02}. To solve this problem, 
\newcite{DBLP:journals/corr/RosenfeldT17} proposed an incremental learning Deep-Adaption-Network that 
constrains newly learned filters to be linear combinations of existing ones.

To the best of our knowledge, for (T)ACSA task, few
researches concerned with 
incremental learning in new categories. In this paper, 
we proposed a (T)ACSA incremental learning task and the \textbf{CNE}-net model to solve this problem in a multi-task learning approach with a shared encoder and shared decoders. We also apply category name for task discrimination.

\section{Datasets}
This section describes the benchmark datasets we used to evaluate our 
model, 
the incremental learning task definition, the methodology to 
prepare the incremental learning dataset, and the evaluation metric.

\label{Datasets}
\subsection{Evaluation Benchmark Datasets}
\label{Evaluation Benchmark Datasets}
We evaluated the performance of the \textbf{CNE}-net model on two benchmark 
datasets, i.e., ACSA task on 
SemEval-2014 Task4 
\cite{DBLP:conf/semeval/PontikiGPPAM14} and TACSA task on SentiHood \cite{DBLP:conf/coling/SaeidiBLR16}.

The \textbf{ACSA task} was evaluated on SemEval-2014 Task4, a 
dataset on restaurant reviews. Our 
model provides a joint solution for sub-task 3 (Aspect Category Detection) and 
sub-task 4 (Aspect Category Sentiment Analysis).
The sentiment polarities are $y \in Y=$  
\{positive, neutral, negative, conflict and none\}, and the categories are $a \in 
A=$ 
\{\textit{food}, \textit{service,} \textit{price}, 
\textit{ambience} and \textit{anecdotes/miscellaneous}\}. The conflict label 
indicates both positive and negative sentiment is expressed in one 
category \cite{DBLP:conf/semeval/PontikiGPPAM14}.

The \textbf{TACSA task} was evaluated on the Sentihood dataset, which describes locations or neighborhoods of London and was collected 
from question answering platform of Yahoo. 
The sentiment 
polarities are $y \in Y=$  \{positive, negative and none\}, the targets are $t 
\in 
T=$  \{Location1, and Location2\}, 
and the aspect categories are $a \in A=$ \{\textit{general}, \textit{price}, 
\textit{transit-location}, 
and \textit{safety}\}.

\subsection{Evaluation Transfer Learning Datasets}
\label{Evaluation Transfer Learning Datasets}

\begin{table*}[!t]
	\centering
	\scalebox{0.9}
	{
		\begin{tabular}{|c|c|}
			\hline
			origin ACSA sample&
			\makecell[c]{\{``text": ``The only thing more wonderful than the food 
				is the service.",\\
				"sentiment": \{``food": ``Positive", ``service": ``Positive", ``price": 
				None,\\
				``ambience": None, ``anecdotes/miscellaneous": None
				\} 
				\}}
			\\
			\cline{1-2}
			ACSA Sample-Source&
			\makecell[c]{\{``text": ``The only thing more wonderful than the food 
				is the service.",\\
				"sentiment": \{``food": ``Positive", 
				``price": 
				None,\\
				``ambience": None, ``anecdotes/miscellaneous": None
				\} 
				\}}
			\\
			\cline{1-2}
			ACSA Sample-Target&
			\makecell[c]{\{``text": ``The only thing more wonderful than the food 
				is the service.",\\
				"sentiment": \{``service": ``Positive"
				\} 
				\}}
			\\
			\cline{1-2}
		\end{tabular}
	}
	\caption{\label{tab:table_add1} An example for generating ACSA incremental 
		learning task.}
\end{table*}
Besides evaluating the model on existing (T)ACSA tasks, we also proposed 
incremental learning tasks for (T)ACSA\footnote{The dataset can be found at \url{https://github.com/flak300S/emnlp2020_CNE-net}.} in new category 
based on SemEval-2014 Task4 and Sentihood dataset, respectively. 

Firstly, we split the categories into source categories and target categories. 
For ACSA task, 
the source 
categories are \{\textit{food}, \textit{price}, \textit{ambience} and 
\textit{anecdotes/miscellaneous}\}, while the target category is 
\{\textit{service}\}. For TACSA task, the source categories are 
\{\textit{general}, \textit{transit-location}, and \textit{safety}\}, while the 
target category is \{\textit{price}\}. This was considered by the amount of 
data with positive/negative/neutral polarity in this category, as well as the sense of this category for real applications.

Secondly, we prepare training, validation and testing data for incremental 
learning task by independently splitting the origin training data, validation data and test data 
 into source-category data 
(\textbf{Sample-Source}) containing label
only in source categories and target-category data (\textbf{Sample-Target}) 
with 
target-category label only. For example, as shown in Table \ref{tab:table_add1}, 
in ACSA task, the 
origin labels 
\{\textit{food}: positive, \textit{service}:positive, 
\textit{price}:none, \textit{ambience}:none, 
\textit{anecdotes/miscellaneous}:none\} were transformed to 
\{\textit{food}: positive, \textit{price}:none, 
\textit{ambience}:none, \textit{anecdotes/miscellaneous}:none\} in 
Sample-Source and 
\{\textit{service}:positive\} in Sample-Target. The input sentences were kept 
the same as origin dataset.
For other researches to investigate the influence of target-category training data amount  quantitatively, we also created incremental learning data by combining all the 
Sample-Source 
and sampled Sample-Target. The sampling rate is a range from 
0.0 to 1.0.

In this paper, the ACSA incremental learning dataset is created from SemEval14-Task ACSA dataset, and it is called SemEval14-Task-\textit{inc}. The TACSA incremental learning dataset is created from Sentihood TACSA dataset, and it is called Sentihood-\textit{inc}.

\subsection{Evaluation Metrics}
\label{Evaluation Metrics} 
We evaluated the aspect category extraction (to determine whether the sentiment is none for each category) and sentiment analysis (to predict the sentiment polarity) on the two datasets. For aspect category extraction evaluation, we applied the probability $1-p$ as the not none probability for each category, where \textit{p} is the probability of the ``none" class in this category. The evaluation metric is 
the same as \newcite{DBLP:conf/naacl/SunHQ19}.
For the origin SemEval-14 Task4 dataset, we use 
Micro-F1 for category extraction evaluation and accuracy for sentiment analysis evaluation. For the  origin Sentihood dataset, we use Macro-F1, strict 
accuracy, and area-under-curve(AUC) 
for category extraction evaluation while use AUC, and strict 
accuracy for sentiment analysis evaluation. When evaluating 
the incremental learning task, we use the F1 metric (Micro-F1 for SemEval-14 and Macro-F1 for Sentihood) for category extraction and accuracy for sentiment analysis.

\section{Approach}

\begin{figure*}[!t]
	\centering
	\includegraphics[width=1.0\textwidth]{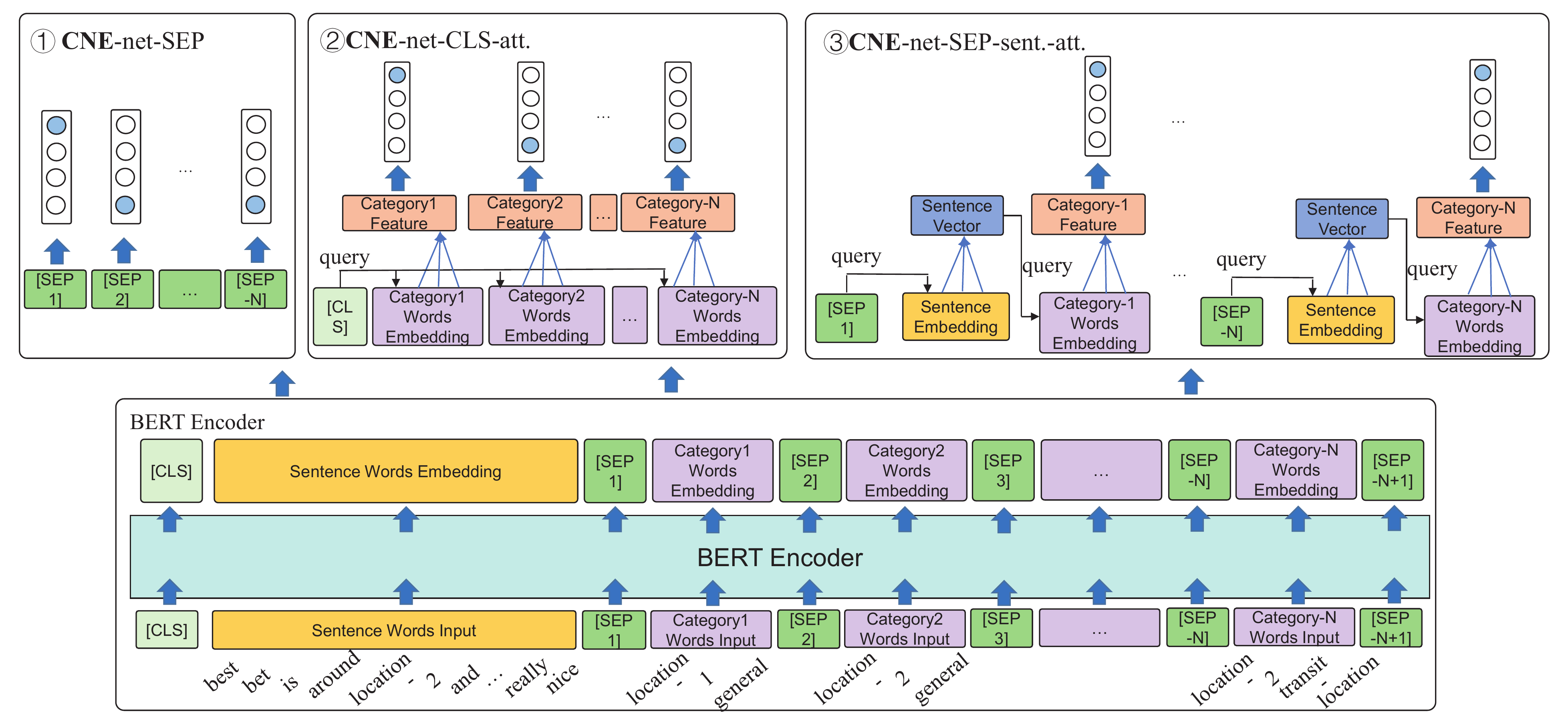}
	\caption{\textbf{CNE}-net model architecture}
	\label{fig:figure1}
\end{figure*}

In this section, we describe the architecture of \textbf{CNE}-net for 
(T)ACSA task. 
In BERT classification tasks, the typical approach is feeding 
sentence ``[CLS]tokens in sentence[SEP]" into the model, while the token 
``[CLS]" is used as a feature for classification. 
In order to encode category 
names into BERT model, as well as analyze sentiment polarity of all the 
categories simultaneously, we made two significant differences from the original BERT, one on the encoder module and another on the decoder module.

\subsection{Encoder with Category Name Embedding}
In order to get a better category name embedding, as well as to make it 
feasible for incremental learning cross categories, the category 
names are encoded into the model, along with the origin sentence like ``[CLS] 
sentence words input [SEP] category1 input 
[SEP] category2 input [SEP]...[SEP] categoryN input[SEP]", as shown in the BERT encoder module in Figure \ref{fig:figure1}. In 
 ACSA task, the category names are 
``\{food, service, price, ambiance, and anecdotes/miscellaneous\}", while in 
 TACSA task, the category names are ``\{location-1 general, 
location-1 price, location-1 transit-location, location-1 safety, location-2 
general, location-2 price, location-2 transit-location, and location-2 
safety\}". 

We mark output states of the BERT encoder as follows: the 
hidden state of 
[CLS] $\vec{h}_{[CLS]}\in \mathbb{R}^{d}$, the hidden states of 
words in origin sentences
$\textbf{H}_{sent} \in \mathbb{R}^{L_{sent}\times d}$, the hidden states of 
separators $\textbf{H}_{[SEP]} \in \mathbb{R}^{n_{cat}\times d}$, and the 
hidden states of category words $\textbf{H}_{cat-i} \in 
\mathbb{R}^{L_{cat-i}\times d}$ for the $i$-th category ($0<i\le n_{cat}$), 
where $L_{sent}$ is the length of the input sentence, $d$ is the dimension of 
hidden states, $n_{cat}$ is the number of categories feed into the 
model, and $L_{cat-i}$ is the length of the $i$-th category input words.

\subsection{Multi-Task Decoders}
We proposed three types of decoder for (T)ACSA task, as shown in Figure 
\ref{fig:figure1}\textcircled{1},\textcircled{2} and \textcircled{3}.
These decoders are multi-label classifiers, which apply a softmax classifier for sentiment analysis in each category.

\textbf{Type 1}, \textbf{CNE}-net-SEP, as shown in 
Figure \ref{fig:figure1}\textcircled{1}, the 
separator token $\vec{h}_{[SEP-i]}$ is applied as feature representation for 
sentiment 
polarity analysis in each category directly. The probability for each polarity in category 
\textit{i} is calculated as follows where $\vec{h}=\vec{h}_{[SEP-i]}$:
\begin{align}
\vec{f}_i = \textbf{W}_i\cdot \vec{h} + \vec{b_i};
\vec{p}_i = softmax(\vec{f}_i)\label{eq:1}
\end{align}
where $\vec{f}_i \in \mathbb{R}^{s}$ is the output logits for category \textit{i}, $\vec{p}_i \in \mathbb{R}^{s}$ is the output probability for 
category \textit{i}, $ \textbf{W}_i \in \mathbb{R}^{d\times s} 
$ and  $ \vec{b_i} \in 
\mathbb{R}^{s}$ are 
randomly initialized parameters to be trained, and $s$ is the number 
of sentiment classes. 
$s=5$ for \{positive, neutral, negative, conflict and none\} in SemEval14-Task4, 
while $s=3$ for \{positive, negative and none\} in Sentihood dataset.
In our approach, $\textbf{W}_1=\textbf{W}_2=...=\textbf{W}_{n_{cat}}$ and 
$\vec{b}_1=\vec{b}_2=...=\vec{b}_{n_{cat}}$.

\textbf{Type 2}, \textbf{CNE}-net-CLS-att., in order to get content-aware 
category embedding vector, we 
applied attention mechanism with $\vec{h}_{[CLS]}$ serves as query vector, and 
$\textbf{H}_{cat-i}$ serves as both key and value matrix, as shown in 
Figure \ref{fig:figure1}\textcircled{2}. The category embedding vector 
$\vec{e}_{cat-i}$ for the \textit{i}-th category is as follows:
\begin{align}
\vec{e}_{cat_i} = 
softmax(\vec{h}_{[CLS]} \cdot \textbf{H}_{cat-i})\cdot 
\textbf{H}_{cat-i}\label{eq:2}
\end{align}
The probability for category 
\textit{i} in type 2 is calculated following equation(\ref{eq:1}) where
$\vec{h}=\vec{e}_{cat_i}$.

\textbf{Type 3}, \textbf{CNE}-net-SEP-sent.-att. applied attention 
mechanism for both sentence embedding and 
category name embedding. As it is shown in 
Figure \ref{fig:figure1}\textcircled{3}. Firstly, sentence vector 
correlated with the \textit{i}-th category is calculated 
by attention with separator embedding $\vec{h}_{[SEP-i]}$ serving as query, and 
sentence embedding $\textbf{H}_{sent}$ serving as key and value matrix. Sentence 
vector 
$\vec{h}_{sent-i}$
correlated with the \textit{i}-th category is as follows:
\begin{align}
\vec{h}_{sent-i} = 
softmax(\vec{h}_{[SEP-i]} \cdot \textbf{H}_{sent})\cdot 
\textbf{H}_{sent}\label{eq:4}
\end{align}
Secondly, similar to that in type 2, the category embedding vector 
$\vec{e}_{cat-i}$ for the \textit{i}-th category calculated by attention mechanism is as follows:
\begin{align}
\vec{e}_{cat_i} = 
softmax(\vec{h}_{sent-i} \cdot \textbf{H}_{cat-i})\cdot 
\textbf{H}_{cat-i}\label{eq:5}
\end{align}
The probability for for category 
\textit{i} in type 3 is calculated 
following equation(\ref{eq:1}) where
$\vec{h}=\vec{e}_{cat_i}$.

\subsection{Model Training}
\label{Model Training}
The \textbf{CNE}-net multi-task framework was trained in an end-to-end way by 
minimizing the sum of cross-entropy loss of all the categories. We 
employed $\textit{L}_2$ regularization to ease over-fitting. The 
loss function is given as follows:
\begin{align}
\textit{L} & = -\frac{1}{\vert D \vert} \sum_{x,y\in D}\sum_{i=1}^{N} \vec{y}_i\cdot
\log \vec{p}_i(x;\theta)+\frac{\lambda}{2}
\vert\vert\theta\vert\vert_2\label{eq:output_3}
\end{align}
where $ D $ is the training dataset, $ N $ is the number of categories, $ Y $ is the 
sentiment classes $Y = \{\textit{positive}, \textit{neutral}, 
\textit{negative}, \textit{conflict}, \textit{none}\}$ (\textit{neutral} and 
\textit{conflict} is not 
included in TACSA 
task), $ 
\vec{y}_i \in \mathbb{R}^{|Y|} $  is the 
one-hot label vector for the \textit{i}-th category with true label 
marked as 1 and others marked as 0, 
$\vec{p}_i(x;\theta)$  
is the probability for the \textit{i}-th category, and 
$\lambda$ is the 
$L_2$ regularization weight. Besides $L_2$ regularization, we 
also employed dropout and early stopping to ease over-fitting. 

During training incremental learning models, we follow the workflow of the
incremental learning application. We firstly train a source-category model with 
the Sample-Source training data.
Then finetuned the source-category model with Sample-Target training
data to get incremental learning model.

\section{Experiments}
\label{Experiments}

\subsection{Experiment Settings}
\label{Experiment Settings}
The pretrained uncased BERT-base\footnote{https://storage.googleapis.com/bert\_models\\/2018\_10\_18/uncased\_L-12\_H-768\_A-12.zip}
was used as the encoder in \textbf{CNE}-net. The number of Transformer blocks is 12, the number of  self-attention heads is 12, and the hidden layer size in each self-attention head is 64. The total amount of parameters in BERT encoder is about 110M. The dropout ratio is 0.1 during training, the traning epochs is 10, and the learning rate is 5e-5 with a warm-up ratio of 0.25.

\subsection{Compared Methods}
\label{Compared Methods}
We compare the performance of our model with some state-of-the-art models.

For ACSA task:  
\begin{compactitem}
	\item XRCE \cite{DBLP:conf/semeval/BrunPR14}: a hybrid classifier based on  
	linguistic features.
	\item NRC-Canada \cite{DBLP:conf/semeval/KiritchenkoZCM14}: several binary 
	one-vs-all SVM classifiers for this multi-class multi-label 
	classification problem.
	\item AT-LSTM and ATAE-LSTM \cite{DBLP:conf/emnlp/WangHZZ16}: a LSTM 
	attention framework with aspect word embeddings concatenated with 
	sentence word embeddings.
	\item BERT-pair-QA-B \cite{DBLP:conf/naacl/SunHQ19}: a question answering 
	and natural language inference model based on BERT.
	\item Multi-task framework (MTL) \cite{DBLP:conf/emnlp/SchmittSSR18}: a LSTM 
	multi-task learning framework with an individual attention head for each 
	category. To better compare our model with this approach, we 
	changed the encoder to BERT-base.
\end{compactitem}

For TACSA task:  
\begin{compactitem}
	\item LR \cite{DBLP:conf/coling/SaeidiBLR16}: a logistic regression 
	classfier with linguistic features.
	\item LSTM-final \cite{DBLP:conf/coling/SaeidiBLR16}: a BiLSTM encoder with 
	final 
	states served as feature representation.
	\item LSTM+TA+SA \cite{DBLP:conf/aaai/MaPC18}: a BiLSTM encoder with 
	complex target-level and sentence-level attention mechanisms.
	\item SenitcLSTM \cite{DBLP:conf/aaai/MaPC18}: LSTM+TA+SA model upgraded by 
	introducing external knowledge.
	\item Dmu-Entnet \cite{DBLP:conf/naacl/LiuCB18}: model with delayed memory 
	update mechanism to track different targets.
	\item Recurrent Entity Network (REN) \cite{DBLP:conf/ecai/Ye020}: a recurrent entity memory network that employs both word-level information and sentence-level hidden memory for entity state tracking.
\end{compactitem}
In TACSA task, besides these models, we also compared our model with the BERT-pair-QA-B model 
and MTL model mentioned in ACSA comparison methods. 

\subsection{Main Results}
\label{Main Results}

\begin{table*}[!t]
	\centering
	\scalebox{0.9}{
		\begin{tabular}{|c|c|c|c|c|c|c|}
			\hline
			\multirow{2}{*}{Model} & 
			\multicolumn{3}{c|}{Category Extraction}&\multicolumn{3}{c|}{Sentiment 
				Analysis}\\
			\cline{2-7}
			
			\multicolumn{1}{|c|}{} & P & R & F & binary & 3-way 
			& 4-way \\
			\hline
			
			XRCE \cite{DBLP:conf/semeval/BrunPR14} & 
			83.23 &81.37 &	82.29 &	- &	- &	78.1	
			\\
			
			NRC-Canada \cite{DBLP:conf/semeval/KiritchenkoZCM14} & 
			91.04 & 86.24 &88.58 &	- &	- &	82.9	
			\\
			
			AT-LSTM \cite{DBLP:conf/emnlp/WangHZZ16} & 
			- & - &	- &	89.6 &	83.1 &	-	
			\\
			
			ATAE-LSTM \cite{DBLP:conf/emnlp/WangHZZ16} & 
			- & - &	- &	89.9 &	84.0 &	-	
			\\
			
			QA-B \cite{DBLP:conf/naacl/SunHQ19} & 
			93.04 & 89.95 &	91.47 &	95.6 &	89.9 &	85.9	
			\\
			
			MTL & 
			91.87 & 90.44 &	91.15 &	95.0 &	88.8 &	85.3	
			\\
			\cline{1-7}
			
			\textbf{CNE}-net-SEP (ours)& 
			92.26 & 90.73 &	91.49 &	95.8 &	90.2 &	86.3	
			\\
			
			\textbf{CNE}-net-CLS-att. (ours)& 
			93.37 & 90.93 &	91.98 &	96.1 &	91.0 &	87.0	
			\\
			
			\textbf{CNE}-net-SEP-sent.-att. (ours)&93.76	
			&90.83	
			&\textbf{92.27}	&\textbf{96.4}	&\textbf{91.3}	&\textbf{87.1}	\\
			\hline
	\end{tabular}}
	\caption{\label{tab:table2} Performance on SemEval-14 Task4, ACSA task. 
		(``-" means not reported.)}
\end{table*}

\begin{table*}[!t]
	\centering
	\scalebox{0.9}{
		\begin{tabular}{|c|c|c|c|c|c|}
			\hline
			\multirow{2}{*}{Model} & 
			\multicolumn{3}{c|}{Category 
				Extraction}&\multicolumn{2}{c|}{Sentiment 
				Analysis}\\
			\cline{2-6}
			
			\multicolumn{1}{|c|}{} & \textit{Acc.} & $F_1$ & AUC & 
			\textit{Acc.} & AUC  \\
			\hline
			
			LR \cite{DBLP:conf/coling/SaeidiBLR16} & 
			- &39.3 &	92.4 &	87.5 &	90.5 	
			\\
			
			LSTM-final \cite{DBLP:conf/coling/SaeidiBLR16} & 
			- & 68.9 &89.8 &	82.0 &	85.4 	
			\\
			
			LSTM+TA+SA \cite{DBLP:conf/aaai/MaPC18} & 
			66.4 & 76.7 & - &	86.8 &	- 
			\\
			
			SenticLSTM \cite{DBLP:conf/aaai/MaPC18} & 
			67.4 & 78.2 &	- &	89.3 &	- 	
			\\
			
			Dmu-Entnet \cite{DBLP:conf/naacl/LiuCB18} & 
			73.5 & 78.5 &	94.4 &	91.0 &	94.8 	
			\\
			
			REN \cite{DBLP:conf/ecai/Ye020} & 
			75.7 & 80.4 &	96.0 &	92.5 &	95.9	
			\\
			
			QA-B \cite{DBLP:conf/naacl/SunHQ19} & 
			79.2 & 87.9 & 97.1 &	93.3 &	97.0 	
			\\
			
			MTL & 
			80.4 & 88.4 &	97.6 &	93.6 &	97.1 	
			\\
			\cline{1-6}
			
			\textbf{CNE}-net-SEP (ours)& 
			80.2 & 88.1 &	97.6 &	93.4 &	97.3 	
			\\
			
			\textbf{CNE}-net-CLS-att. (ours)& 
			80.4 & 88.8 &	97.8 &	93.8	&97.4
			\\
			
			\textbf{CNE}-net-SEP-sent.-att. 
			(ours)&\textbf{80.8}	
			&\textbf{89.4}	
			&\textbf{97.9}&	\textbf{94.0} &	\textbf{97.5} 	
			\\
			\hline
	\end{tabular}}
	\caption{\label{tab:table3} Performance on Sentihood, TACSA task. 
		(``-" means not reported.)}
\end{table*}

The performances of compared methods and three types of \textbf{CNE}-net are shown in 
Table \ref{tab:table2} (ACSA task) and 
Table \ref{tab:table3} 
(TACSA task). All the models with BERT encoder 
(QA-B, MTL and our \textbf{CNE}-net)
achieved better performance compared with models without BERT encoder (XRCE, NCR-Canada, AT-LSTM, ATAE-LSTM, SenitcLSTM, Dmu\_entnet, and REN). 
Our \textbf{CNE}-net performs better compared with QA-B and MTL 
framework in both ACSA and TACSA tasks. QA-B is a single-task approach, which each category is trained independently. Our 
\textbf{CNE}-net is a multi-task learning framework. It performs better 
than QA-B by 
using shared semantic features and sentiment labels in all the categories.
\textbf{CNE}-net also performs better 
compared with the MTL model since it 
encodes the category names as additional features to generate the 
representation of each category.

Our \textbf{CNE}-net-SEP-sent.-att. model achieves state-of-the-art 
on all the evaluation metrics in both SemEval14-Task4 and 
Sentihood dataset. The improved 
extraction $F_1$ is 0.0080 
in the SemEval14-Task4 (increased from 0.9147 in QA-B to 0.9227 in \textbf{CNE}-net-SEP-sent.att.), while it is 0.010 in the Sentihood dataset (increased from 0.884 in MTL to 0.894 in \textbf{CNE}-net-SEP-sent.att.). The accuracy metrics for sentiment analysis in the
SemEval14-Task4 are binary, 3-way and 4way, which refers to accuracy with 
positive/negative (binary), positive/neutral/negative (3-way) and 
positive/neutral/negative/conflict (4-way). The improvement of 
sentiment classification accuracy is 0.012 in 
SemEval14-Task4 (4-way setting, increased from 0.859 in QA-B to 0.871 in \textbf{CNE}-net-SEP-sent.att.), while is 0.004 in the Sentihood dataset (increased from 0.971 in MTL to 0.975 in \textbf{CNE}-net-SEP-sent.att.). 

\begin{table*}[!t]
	\centering
	\scalebox{0.9}{
		\begin{tabular}{|c|c|c|c|c|c|c|c|c|}
			\hline
			\multirow{2}{*}{Model} & 
			\multicolumn{4}{c|}{SemEval14-Task4-\textit{inc}}&\multicolumn{4}{c|}{Sentihood-\textit{inc}}\\
			\cline{2-9}
			
			&\multicolumn{2}{c|}{$extra.$}&\multicolumn{2}{c|}{$senti.$}&
			\multicolumn{2}{c|}{$extra.$}&\multicolumn{2}{c|}{$senti.$}\\
			\cline{2-9}
			
			& $mix.$ & 
			$incre.$ & $mix.$ & 
			$incre.$ &$mix.$ & 
			$incre.$ &$mix.$ & 
			$incre.$  \\
			\hline
			
			AE-LSTM & 85.3	&85.0	&85.2	&85.9	&86.3	&86.5&	84.4&	84.5  	
			\\
			
			ATAE-LSTM & 85.6&	85.2	&85.4	&86.0	&86.6	&86.9	&84.6	&84.7
			\\
			
			Dmu-Entnet & - &- &	-  &-	
			& 87.9&	88.0&85.4&	85.8 	
			\\
			
			QA-B &92.2&92.5&91.9&92.0&93.7&93.6&90.6&91.0	
			\\
			
			MTL &92.5	&92.6	&92.4	&92.5&	93.8&	93.7&	90.8&	91.4 	
			\\
			
			\cline{1-9}
			\textbf{CNE}-SEP(ours) &92.9&	92.7	&92.5	&92.8	&94.5	&94.8	&91.2	&91.6 	
			\\
			
			\textbf{CNE}-net-CLS-sent.(ours) &93.0&92.8&92.7&93.0&94.8&95.0&91.6&91.7	
			\\
			
			\textbf{CNE}-net-SEP-sent.-att. 
			(ours) &\textbf{93.6} &	\textbf{93.7} &	\textbf{93.0} &	\textbf{93.2} & 
			\textbf{95.2}& \textbf{95.4}&\textbf{91.9} &\textbf{92.0}	
			\\
			
			\hline
	\end{tabular}}
	\caption{\label{tab:table4} Extraction $F_1$ and sentiment accuracy in target category of incremental 
		learning.}
\end{table*}

\begin{table*}[!t]
	\centering
	\scalebox{0.9}{
		\begin{tabular}{|c|c|c|c|c|c|c|c|c|}
			\hline
			\multirow{2}{*}{Model} & 
			\multicolumn{4}{c|}{SemEval14-Task4-\textit{inc}}&\multicolumn{4}{c|}{Sentihood-\textit{inc}}\\
			\cline{2-9}
			
			&\multicolumn{2}{c|}{$extra.$}&\multicolumn{2}{c|}{$senti.$}&
			\multicolumn{2}{c|}{$extra.$}&\multicolumn{2}{c|}{$senti.$}\\
			\cline{2-9}
			
			& $mix.$ & 
			$incre.$ & $mix.$ & 
			$incre.$ &$mix.$ & 
			$incre.$ &$mix.$ & 
			$incre.$  \\
			\hline
			
			AE-LSTM & 83.6& 83.4&78.3 &77.9 & 82.3&81.5 &85.1 &84.0  	
			\\
			
			ATAE-LSTM & 83.7& 83.5&78.7 &78.0 &82.6 &81.6 &85.6 &85.0	  	
			\\
			
			Dmu-Entnet & - &- &	-  &-	&83.2 &82.3 &85.8 &85.2  	
			\\
			
			QA-B & 90.0& 89.2&84.4 &83.5	& 85.2& 84.2& 91.7&90.7 	
			\\
			
			MTL & 89.8&69.8$\downarrow$ &84.5 & 82.3& 87.0& 75.7$\downarrow$&92.2 & 91.0	
			\\
			
			\cline{1-9}
			\textbf{CNE}-SEP(ours) & 90.9&90.1 &84.8 &84.5 &87.2 &85.8 &92.6 &91.6 	
			\\
			
			\textbf{CNE}-net-CLS-sent.(ours) &91.2 &91.1 &85.4 &85.0 &87.5 &86.1 &93.0 &91.9 	
			\\
			
			\textbf{CNE}-net-SEP-sent.-att. 
			(ours) & \textbf{91.6}& \textbf{91.3}& \textbf{85.5}& \textbf{85.4}& \textbf{87.7}& \textbf{86.3}& \textbf{93.2}&	\textbf{92.3}
			\\
			\hline
	\end{tabular}}
	\caption{\label{tab:table7} Extraction $F_1$ and sentiment accuracy in source categories of incremental learning.}
\end{table*}

\textbf{CNE}-net-SEP 
uses [SEP] as a feature representation for sentiment classification. It 
performs the poorest among all three types of \textbf{CNE}-net since 
representation 
from only [SEP] token does not make full use of sentence information and 
category information. \textbf{CNE}-net-CLS-att. uses [CLS] as 
sentence representation and 
applies attention mechanism to build the relationship between sentence 
representation 
and the category name hidden states to get sentiment classification feature and achieve better performance. The 
\textbf{CNE}-net-SEP-sent.-att. uses attention twice. The first one is to 
build category-name-aware sentence embeddings for each category with [SEP] 
as 
query and 
sentence hidden states matrix as key and value, while the second one is to 
apply each category-name-aware sentence embedding to generate category 
representation like what 
we do in \textbf{CNE}-net-CLS-att.. This category-name-aware sentence 
embedding 
and the sentence-aware category embedding makes it perform the best in the 
three types of \textbf{CNE}-net. 

\subsection{Incremental Learning Results}
\label{Incremental Learning Results}

This section describes the performance in the incremental learning task. 
We trained the model following incremental learning workflow, as mentioned in 
section \ref{Model Training}. We compared the results between mix-training (short as $mix.$) (mixing Sample-Source and Sample-Target) and incremental learning (short as $incre.$), for both extraction $F_1$ and sentiment accuracy.

Firstly, we compare the performance in target category, i.e. aspect category 
extraction $F_1$ (short as $extra.$) and sentiment analysis accuracy (short as 
$senti.$) from mix-training process  and incremental learning. 
As the target category performance
shown in Table \ref{tab:table4}, there is no significant difference between mix-training and incremental learning for both aspect extraction and sentiment analysis.
For example, in SemEval14-Task-\textit{inc}, the extraction $F_1$ and sentiment accuracy of \textbf{CNE}-net-SEP-sent.-att. are 0.936 and 0.930 respectively in mix-training, while they are 0.937 and  0.932 respectively in incremental learning. In Sentihood-\textit{inc}, the extraction $F_1$ and sentiment accuracy of \textbf{CNE}-net-SEP-sent.-att. are 0.952 and 0.919 respectively in mix-training, while they are 0.954 and 0.920 respectively in incremental learning.
This indicates incremental learning does not decrease the performance in the target category. Our
\textbf{CNE}-net-SEP-sent.-att. performs the best in all the models. 

\begin{table*}[!h]
	\centering
	\scalebox{0.9}{
		\begin{tabular}{|c|c|c|c|c|c|c|c|c|}
			\hline
			\multirow{2}{*}{\textbf{CNE}-net-SEP-sent.-att.} & 
			\multicolumn{4}{c|}{SemEval14-Task4-\textit{inc}}&\multicolumn{4}{c|}{Sentihood-\textit{inc}}\\
			\cline{2-9}
			
			&\multicolumn{2}{c|}{Source Categories}&\multicolumn{2}{c|}{Target Category}&
			\multicolumn{2}{c|}{Source Categories}&\multicolumn{2}{c|}{Target Category}\\
			\cline{2-9}
			
			& $extra.$ & 
			$senti.$ & $extra.$ & 
			$senti.$ &$extra.$ & 
			$senti.$ &$extra.$ & 
			$senti.$  \\
			\hline
			
			shared decoder 
			&	91.3 &	85.4&93.7 &	93.2 
			& 86.3& 92.3& 95.4&92.0 
			
			\\
			
			unshared decoder
			 &84.2$\downarrow$ &84.0 & 93.4&93.0
			 & 79.6$\downarrow$& 91.5& 94.9&91.6	
			
			\\
			\hline
	\end{tabular}}
	\caption{\label{tab:table6} Extraction $F_1$ and sentiment accuracy after incremental learning of  \textbf{CNE}-net-SEP-sent.-att. with shared and unshared decoder.}
\end{table*}

Secondly, we compare aspect extraction and sentiment analysis 
performance in source categories after incremental learning, since both 
source categories and target categories requires high accuracy. The 
extraction $F_1$ and sentiment accuracy
of source categories after the incremental learning process as well as in the mix-training process are shown in Table \ref{tab:table7}. There is no significant difference 
in sentiment accuracy of source categories after training with incremental 
learning data. For example, in SemEval14-Task-\textit{inc}, sentiment accuracy of \textbf{CNE}-net-SEP-sent.-att. is 0.855 in mix-training, while it is 0.854 in incremental learning. This is probably because of the similar sentiment features between categories, in which the fine-tuning process does not make a great difference. 

However, for category extraction, MTL suffers from
catastrophic forgetting after fine-tuning. In 
SemEval14-Task4-\textit{inc}, extraction $F_1$ of MTL model of  
source categories decreases from 0.898 in mix-training to 0.698 after incremental learning, while in Sentihood-\textit{inc}, $F_1$ metric of MTL model of source categories decreases from 0.870 in mix-training to 0.757 after incremental learning. Fortunately, the 
QA-B model, as well as our \textbf{CNE}-nets, suffer less from this problem. In SemEval14-Task4-\textit{inc}, extraction $F_1$ metric of 
\textbf{CNE}-SEP-sent.-att. is 0.913 in source categories after 
fine-tuning, while it is 0.916 in mix-training. In Sentihood-\textit{inc}, extraction $F_1$ of 
\textbf{CNE}-SEP-sent.-att. is 0.863 in source categories after 
fine-tuning, while it is 0.877 in mix-training.

\subsection{Discussion}
\label{Discussion}

We have confirmed the effectiveness of \textbf{CNE}-nets for (T)ACSA tasks and (T)ACSA incremental learning tasks. However, there remains a question, why our model suffers less from catastrophic forgetting in incremental learning? 

To answer this question, we compare the incremental learning performance of our \textbf{CNE}-net-SEP-sent.-att. with a similar model but the decoders in each category are unshared with $\textbf{W}_1\not=\textbf{W}_2\not=...\not=\textbf{W}_{n_{cat}}$ and
$\vec{b}_1\not=\vec{b}_2\not=...\not=\vec{b}_{n_{cat}}$
(\textbf{CNE}-net-SEP-sent.-att.-unshared) in equation (1) and the results are shown in Table  \ref{tab:table6}. There is no significant difference in target category between the model with shared decoders and the model with unshared decoders, indicating both shared and unshared model is able to get enough feature for category extraction and sentiment analysis in target category.
However, it is more important that, in \textbf{CNE}-net-SEP-sent.-att.-unshared, the extraction $F_1$ suffers from a sudden decrease. In SemEval14-Task4-\textit{inc}, extraction $F_1$ decreases from 0.913 with shared decoder to 0.842 with unshared decoder, while in Sentihood-\textit{inc}, extraction $F_1$ decreases from 0.863 with shared decoder to 0.796 with unshared decoder. 

We believe the decreased extraction $F_1$ in source categories is due to the unshared decoders for each task, which results in only shared encoder and target-category decoders are fine-tuned during the fine-tuning process. In contrast, the decoder of source categories remains unchanged. The finetuned encoder and original source-category decoder is the reason for the  catastrophic forgetting problem in the category extraction evaluation. In our shared decoder approach, both encoders and decoders are shared and fine-tuned to weaken the catastrophic forgetting problem. 

\section{Conclusion}

In this paper, in order to make multi-task learning feasible for incremental learning, 
we proposed \textbf{CNE}-net with different attention mechanisms. The category 
name features and the multi-task learning structure help the model 
achieve state-of-the-art on ACSA and TACSA tasks. Furthermore, 
the shared encoder and decoder layers weaken catastrophic forgetting in the incremental learning task. 
We proposed a task for (T)ACSA incremental learning and achieved the best 
performance with \textbf{CNE}-net compared with other strong baselines. 
Further research may be concerned with zero-shot learning on new categories.

\bibliography{emnlp2020}
\bibliographystyle{acl_natbib}
\end{document}